\documentclass{article}

\usepackage{PRIMEarxiv}
\usepackage{amsmath}
\usepackage[utf8]{inputenc} 
\usepackage[T1]{fontenc}    
\usepackage{hyperref}       
\usepackage{url}            
\usepackage{booktabs}       
\usepackage{amsfonts}       
\usepackage{nicefrac}       
\usepackage{microtype}      
\usepackage{lipsum}
\usepackage{multirow}
\usepackage{fancyhdr}       
\usepackage{graphicx}       
\graphicspath{{media/}}     
\usepackage{subcaption} 
\title{Tag-Based Annotation for Avatar Face Creation}

\author{
  An Ngo \\
  UC Santa Cruz \\
  \texttt{andingo@ucsc.edu} \\
   \And
  Daniel Phelps \\
  UC Santa Cruz \\
  \texttt{dasphelp@ucsc.edu} \\
  \And
  Derrick Lai \\
  UC Santa Cruz \\
  \texttt{dlai3@ucsc.edu} \\
  \And
  Thanyared Wong \\
  UC Santa Cruz \\
  \texttt{thawong@ucsc.edu} \\
  \And
  Lucas Mathias \\
  UC Santa Cruz \\
  \texttt{lamathia@ucsc.edu} \\
  \And
  Anish Shivamurthy \\
  UC Santa Cruz \\
  \texttt{ashivamu@ucsc.edu}
  \And
  Mustafa Ajmal \\
  UC Santa Cruz \\
  \texttt{muajmal@ucsc.edu}
  \And
  Minghao Liu \\
  UC Santa Cruz \\
  \texttt{mliu40@ucsc.edu} \\
  \And
  James Davis \\
  UC Santa Cruz \\
  \texttt{davis@cs.ucsc.edu} \\
}

\begin{document}
\maketitle

\begin{abstract}
Currently, digital avatars can be created manually using human images as reference. Systems such as Bitmoji are excellent producers of detailed avatar designs, with hundreds of choices for customization. A supervised learning model could be trained to generate avatars automatically, but the hundreds of possible options create difficulty in securing non-noisy data to train a model. As a solution, we train a model to produce avatars from human images using tag-based annotations. This method provides better annotator agreement, leading to less noisy data and higher quality model predictions. Our contribution is an application of tag-based annotation to train a model for avatar face creation. We design tags for 3 different facial facial features offered by Bitmoji, and train a model using tag-based annotation to predict the nose. 
\end{abstract}

\section{Introduction}
The traditional annotation method for training a model to predict avatars would involve human annotators manually matching human face images to their corresponding avatars. This has limitations due to the ambiguity in the process of avatar creation, resulting in low annotator agreement and noisy data. This method creates a dataset with a significant amount of label noise which is ineffective for machine learning.

Alternatively we implement a tag-based annotation method for avatar creation which is much more efficient in minimizing label noise within a data set. A tag based system uses a predetermined set of semantically relevant tags for each feature of the face. Annotators label images with tags, which are then used to train a model. The model predicts tags given a human image, and from those tags, an algorithm can be implemented to convert predicted tags to an avatar.

The difficulty with a tag-based system is that it struggles when tags are not descriptive enough to represent specific features. This is especially prevalent in facial features where the subtleties are harder to express in tags. For example, a tag that describes the width of the nose can be subjective amongst annotators. Eyes also have many small differences in orientation and shape that although are important in avatar creation, are not easily distinguished by unskilled annotators. 

To address this, we incorporate metrics based off of annotator agreement in our tag design process to ensure our tags are agreeable. We iterate on tag design until metrics show tags are agreeable enough amongst researchers to be used for training. Annotators are given reference sheets for each feature to disambiguate the process.

We go through a process of tag design for eyes, noses, and eyebrows and demonstrate a high agreement percentage. A model was trained using images labeled with nose tags to predict noses, but the results are inconclusive as to whether or not tag-based annotation is effective for this specific feature.

\section{Related Works}
\textbf{Tag-based annotation for hair:}
An end-to-end implementation of the tag-based approach for avatar hair generation has been successful \cite{liu2023tag}. The method shows that detailed tags create higher annotator agreement leading to less noise in labeled hair, allowing supervised learning models trained on the data to converge more optimally. In addition, the tags are able to generalize to multiple avatar systems, namely Google Cartoonset, Metahuman, and NovelAI \cite{Cole_Mosseri_Krishnan_Sarna_Maschinot_Freeman_Fuman, unrealengineMetaHumanRealistic, novelaiNovelAIStoryteller}. A search algorithm using a database of Bitmoji avatars was used to display the results from a model trained on hair \cite{bitmoji}. The Fairface dataset was used in this study for annotation and training because it is racially diverse.

We use the Fairface dataset as well in our study for consistency and diversity in facial features among the faces. We apply the iterative tag design process described in this paper to find the best tags for features. Models are trained to predict each tag separately. Our model results are represented by Bitmojis primarily \cite{bitmoji}. Our research focuses on applying tag-based annotation for the features of the face rather than the hair. Characterizing facial features with tags compared to hair is inherently more difficult. As a result, our tag design process is more thorough as we incorporate reference sheets and only move forward with training when our tags reach sufficient accuracy on our metrics.

\textbf{Geometric Feature Extraction:}
Facial features can be extracted geometrically, then classified using a machine learning algorithm. One approach is placing a grid over the face to extract features, then concatenating the grids to classify a feature \cite{rivera2012local, moore2011local}. Murugappan and Mutawa used triangles created by markers placed on a subject's face to derive areas, circumferences, and inscribed circle areas that correspond to 6 basic emotions \cite{murugappan2021facial}. A machine learning algorithm classifies the emotion of the subject based on these changing parameters. This approach was shown to work well for classifying emotions. However, in the context of classifying facial feature attributes with a dataset like Fairface, differences in image quality, orientation, and lighting would cause huge noise within a triangle or grid system’s measurements. As a result, this approach cannot be applied to extract features from faces.

\textbf{Face datasets:}
There exist many datasets of human faces \cite{cao2018vggface2, 8272731, guo2016ms}. In addition, face datasets with annotations of tags describing race and facial features also exist \cite{liu2015deep, karkkainen2019fairface}.  Terhoorst et. al. investigated annotation with tags on the CelebA dataset and the quality of such annotations \cite{terhorst2021maad}. The quality of the annotations is represented through accuracy, precision, and recall. Tags that are more subjective such as whether or not someone has an oval face are shown to have much lower accuracy than a tag that is more concrete, such as the color of one’s hair. These datasets include some tags that may be generalizable to our Bitmoji system of avatars, but most of the tags are not specific enough to represent the multitudes of options given by Bitmoji to create an accurate avatar.

\section{Methods}

\subsection{Sampling Images}
The Fairface dataset was an ideal option to sample from when compared to other datasets because it is fairly balanced in terms of race, gender, and age. 

For tag design, researchers manually picked out a set of 100 faces from Fairface that were most clear and diverse, in order to iterate on tags and come up with clear definitions for how to categorize facial features \cite{karkkainen2019fairface}. While training machine learning, a larger set of images was needed. To ensure images annotated contained as little noise as possible, images were removed from the dataset using inference from a pre-trained facial detection model \cite{githubGitHubTimeslerfacenetpytorch}. Images where the model detected a face with a confidence score of 1.0 were kept, whereas images with lower scores were discarded. A set of 2,741 images out of a 10,000 image sample from the Fairface training set was created using this method. 

\subsection{Tag Design}
\begin{figure}[!t]
    \centering
    \includegraphics[scale=0.5]{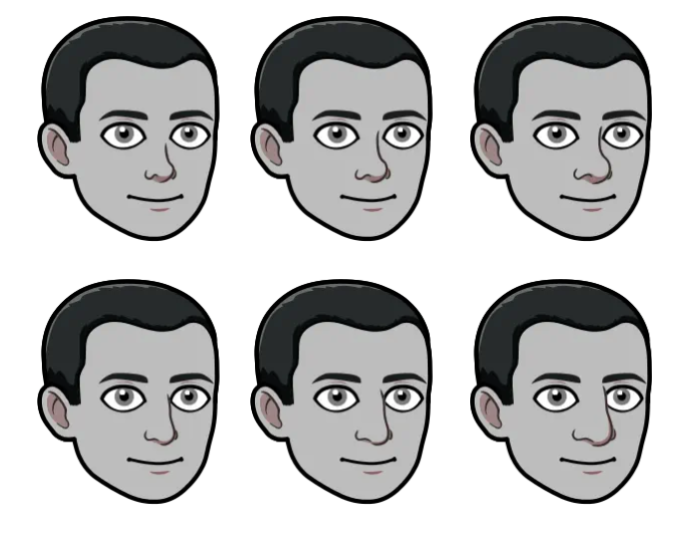}
    \caption{Bitmoji nose progression across two styles. The style of the nose is the same for each row, but the length and width of the nose change. All Bitmoji nose options share this progression. }
    \label{BitmojiProgression}
\end{figure}
\begin{table}[!h]
  \centering
  \begin{tabular}{lll}
    \hline
    \textbf{nose tags} & \textbf{eye tags} & \textbf{eyebrow tags} \\
   \hline
    width narrow & width extra small & thin density \\

    width medium & width small & medium density \\

    width wide & width medium  & bushy density \\

    bridge downturned & width wide & thickness thin \\

    bridge upturned & angle upturned & thickness medium\\

    bridge straight & angle downturned & thickness thick\\

    tip downturned & angle straight & s-shaped \\

    tip upturned & top crease & round shape\\

    tip straight & no top crease & angled shape \\

    no nasal hump & sharp curvature \\

    nasal hump & medium curvature \\

    bridge projection flat & round curvature \\

    bridge projection medium & & \\

    bridge projection tall & & \\

    tip projection flat & &  \\

    tip projection medium & &\\

    tip projection tall & & \\

   \hline \\
\end{tabular}
   \caption{\textbf{Final attributes used for tag-based annotation and training}: Nose tags for tag-based annotation are separated into tag categories: the width of the nose, the style of the nose bridge, the style of the nose tip, the projection of the nose bridge (how far it comes out of the face), the projection of the nose tip, and whether or not there is a nasal hump or not. Eye tags rely much more on the reference sheet as curvature and angle are determined by a geometrical test. The tags are separated into the width of the eye, angle, curvature, and whether or not there is a top crease. Eyebrow tags are separated into density, thickness, and shape. The individual options within each tag category are either a progression in size or changes in style. These options are outlined in the reference sheets in an effort to make them less subjective.}
   \label{tab:tags}
\end{table}
\textbf{Tag Creation:} We analyzed Bitmoji assets and Fairface images, looking for facial features that saw high differentiation amongst people/avatars. It was helpful to understand differences among samples for a given feature as varying along a spectrum. Ranking a tag from "small," to "medium," to "large," for example, gives a tag like "nose projection" a quantifiable value and makes the tag selection process intuitive. We observed that Bitmoji tended to present noses on their avatar in three different sizes with a multitude of styles. Designing the nose tags in a similar way allows tags to match human noses to avatar noses more accurately (Fig. \ref{BitmojiProgression}).

\textbf{Design Iteration:}
Researchers started with initial tag definitions and participated in annotation sessions to evaluate tags and adjust them according to annotator agreement metrics. Each session consisted of 3-4 annotators, tagging a given image batch to evaluate tag performance. This image batch is taken from Fairface dataset, and is randomized for each annotation session to prevent researchers from learning images. 
\subsection{Annotation Metrics}
The creation of descriptive and effective tags to describe images is immensely important in raising annotator agreement and gathering high quality data through crowdsourcing platforms such as Mechanical Turk \cite{liu2023tag}\cite{mechanicalturk}. We focused on this process of tag design by developing an annotation simulator, a tool that would allow us to simulate annotation amongst researchers, testing different tags on different facial features quickly with instantaneous analytics and results. 

Our annotation simulator displays the following metrics:

\medskip \textbf{Tag agreement metric:} 
This metric measures the proportion of people who agree when selecting a tag, or how agreeable that tag is when it appears during annotation.

\begin{equation} \label{eq:1}
    \frac{k}{n},
\end{equation}
\indent{where \textit{k} is the number of annotators who agree and \textit{n} is the number of annotators.} \\

\medskip \textbf{All tags per image agreement metric:} This metric outputs the average agreement of all tags in an image. This metric is useful for finding disagreeable images quickly. 
\begin{equation} \label{eq:2}
    \frac{\sum _{i=1}^{t} x_i}{t},
\end{equation}
\indent{where $x_i$ is the tag agreement percentage for the $i$-th tag in an image as calculated by equation (1), and \textit{t} is the \indent number of agreed upon tags in an image.} \\ \\

\medskip \textbf{Overall tag agreement metric:} This metric measures how agreeable a tag was during an annotation session. It is useful for singling out weak tags.
\begin{equation} \label{eq:3}
    \frac{\sum _{i=1}^{a} y_i}{a}
\end{equation}
\indent{where $y_i$ is the tag's tag agreement for the $i$-th image in the batch as calculated by equation (1). \textit{a} is the number of times \indent the tag had agreement across all images in an annotation session.}
\begin{figure}[!h]
    \centering

    \begin{subfigure}{0.3\textwidth}
        \centering
        \includegraphics[width=\linewidth]{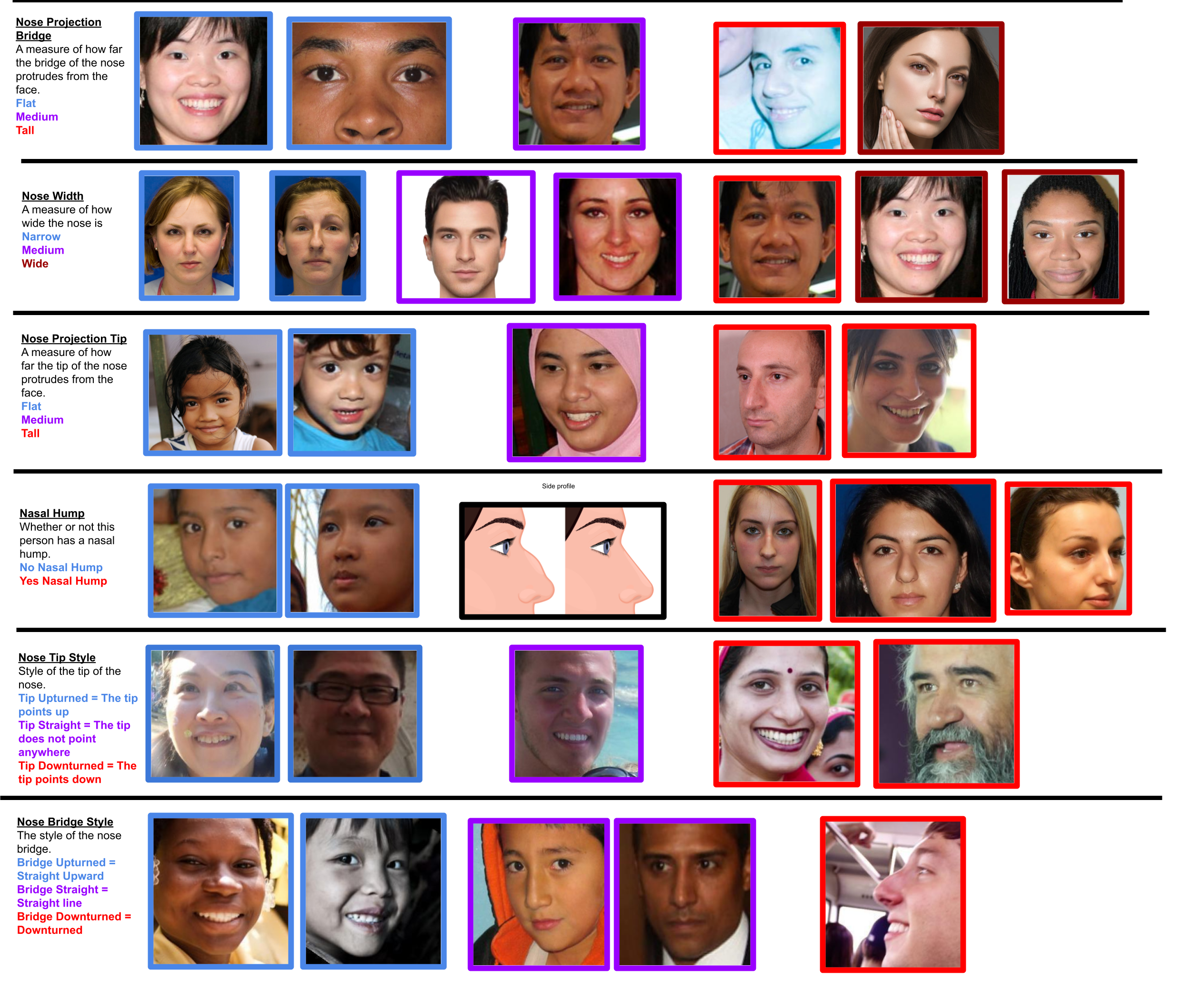}
        \caption{Nose reference Sheet}
        \label{fig:subfig1}
    \end{subfigure}
    \hfill
    \begin{subfigure}{0.3\textwidth}
        \centering
        \includegraphics[width=\linewidth]{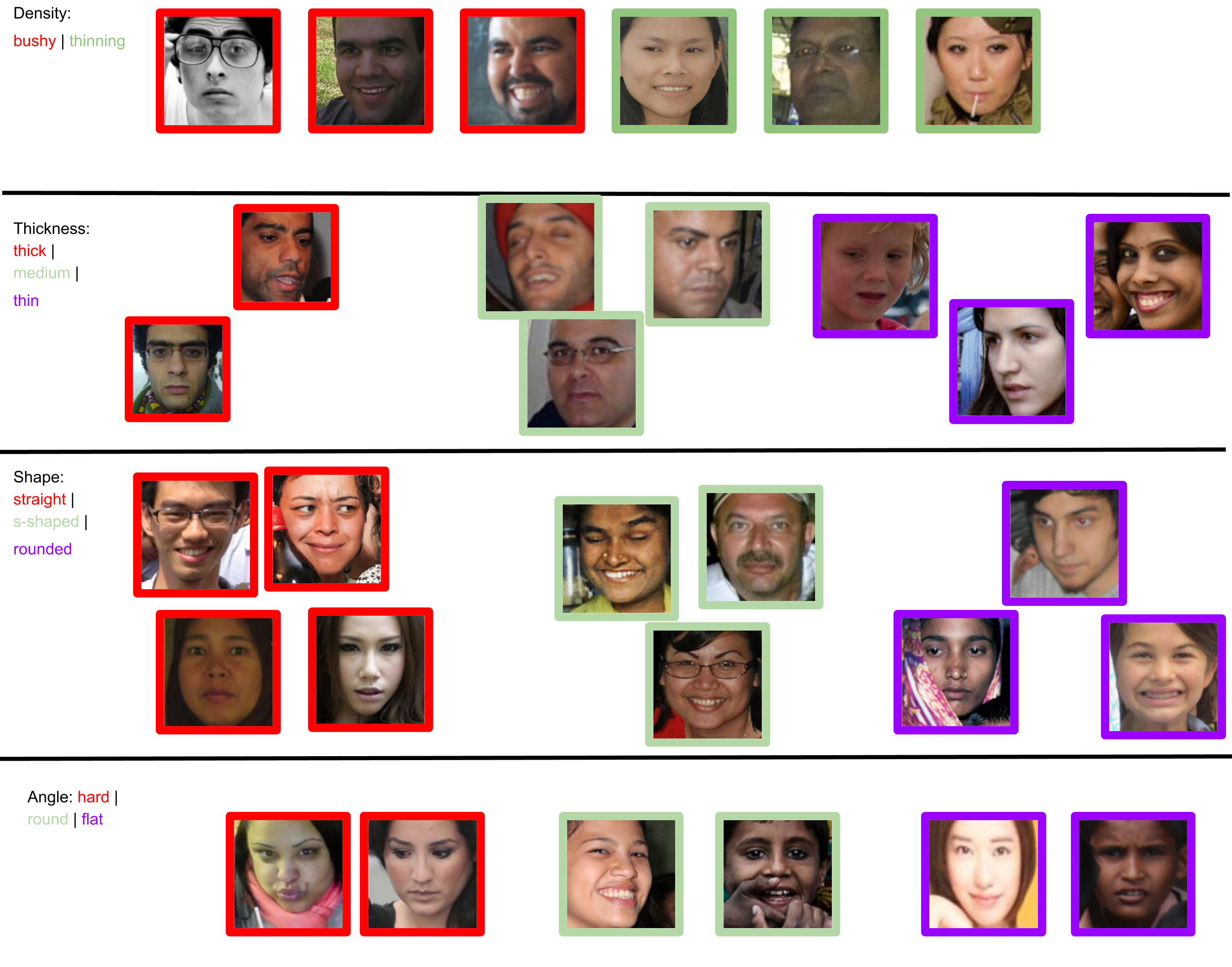}
        \caption{Eyebrow reference sheet}
        \label{fig:subfig2}
    \end{subfigure}
    \hfill
    \begin{subfigure}{0.3\textwidth}
        \centering
        \includegraphics[width=\linewidth]{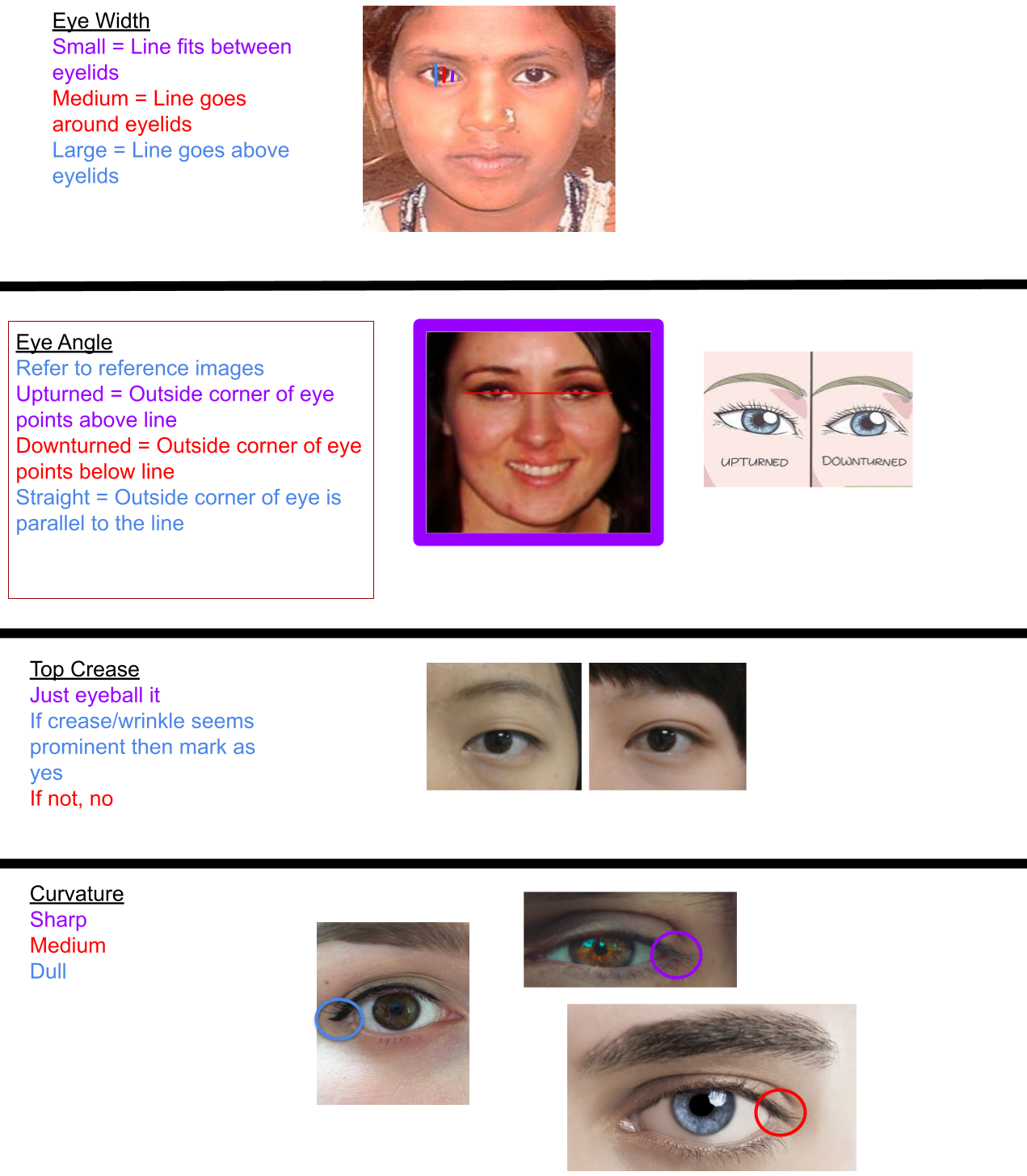}
        \caption{Eye reference sheet}
        \label{fig:subfig3}
    \end{subfigure}

    \caption{Final reference sheets used to tag. Tags along with short descriptions are displayed on the left, and color coded images corresponding to the tags are provided on the right. }
    \label{fig:ref_sheets}
\end{figure}

\subsection{Reference Sheets}
To support researchers in adhering to tag definitions, reference sheets were created consisting of a careful selection of images that exemplify each feature and make their distinctions clear (Fig.\ref{fig:ref_sheets}). Some reference sheets images included markings and labels to clarify tag definitions further.

\subsection{Gathering Labels}

\begin{figure}[t]
    \centering
    \includegraphics[scale=0.45]{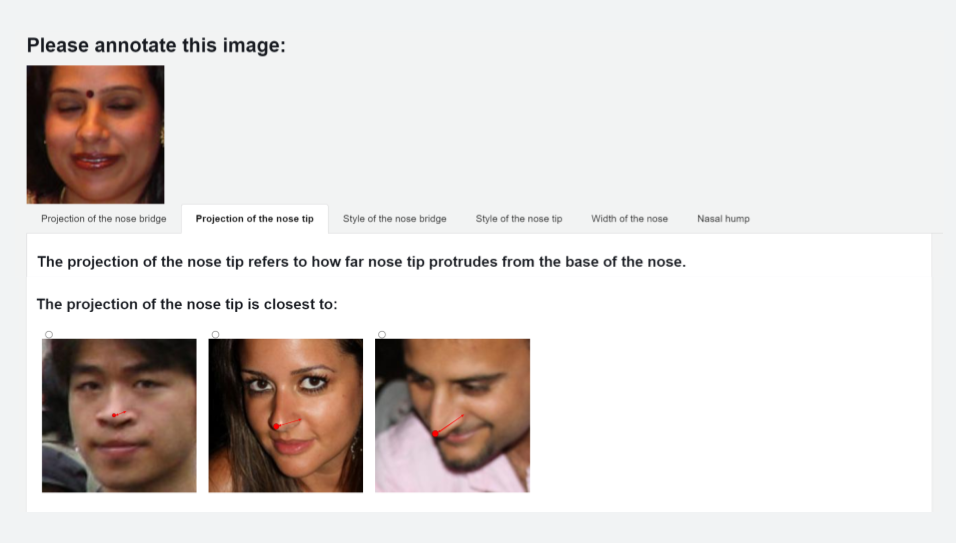}
    \caption{Amazon Mechanical Turk user interface for labeling task. Turkers navigate through tabs consisting of tag categories to select the image that matches closely with a specific tag attribute. The option images map to specific tags. Images were selected for annotation using the process outlined in section 3.1.}
    \label{TurkUI}
\end{figure}
Amazon Turk was used to gather labels for training a model. Tag options within a category are displayed horizontally, giving equal access to each tag without having to scroll up or down. Secondly, reference images are integrated into the MTurk UI so that each selection for a tag is accompanied by an image. Instead of seeing 'nose\_projection\_bridge\_tall,' a MTurk worker sees the reference image for a person with a tall projection bridge with the feature highlighted in the image. The goal of this practice is to make the definition of a tag objective. Target images can be directly compared to references so that 'small,' 'medium,' and 'large,' for instance, becomes easily distinguishable.

\subsection{Machine Learning}
We train 6 Resnet152 models, pretrained on Imagenet weights, predicting nose tags \cite{pytorchModelsPretrained, He_2016_CVPR}. Each category of tags (nose width, nose tip style, etc.) were trained separately in order to view how well the model was learning the separate features. Cross entropy loss was used.

\subsection{Bitmoji Conversion Algorithm}
Given a model that can produce tags from a given image, these tags can be used to map a Bitmoji avatar to the user. Bitmoji conversion from tags requires that Bitmoji assets are tagged individually and stored in a database. A conversion algorithm finds the best matched asset for a given set of tags. We implement this algorithm for our nose tags.

\begin{table}[!ht]
\centering
\begin{tabular}{lllll}
\hline
\textbf{Region} & \textbf{Annotation Tags} & \textbf{\#Options} & \textbf{Weight (1-5)} & \textbf{Type} \\
\hline
\multirow{6}{*}{Nose} & Width & 3 & 5 & Continuous \\
& Nose Tip Projection & 3 & 4 & Continuous \\
& Nose Bridge Projection & 3 & 3 & Continuous \\
& Nasal hump & 2 & 5 & Discrete \\
& Nose Tip Style & 3 & 4 & Continuous \\
& Bridge Style & 3 & 3 & Continuous \\
\hline
\end{tabular} 
\caption{Weighting for Bitmoji conversion algorithm.}
\label{tab:distance}
\end{table}

Best match is defined by concurrence of important features. All features are weighted against each other so that tags like "Nose Width," for example, contributes more weight towards a better match than "Nose Tip Style." The weightings were decided by researchers after an evaluation of which tags were most important to identifying a person's nose. \\


\section{Results}
The numerical values presented in the following tables have been derived from the metrics elaborated upon in Section 3.3.
Tables 3-8, 10-13, 15-17 are calculated using our tag agreement metric, equation \eqref{eq:3}. Tables 9, 14, and 18 are calculated using our image agreement metric, equation \eqref{eq:2}.  

We report reasonable to high annotator agreement numbers on the overall tag agreement metric for most nose, eye, and eyebrow tags, indicating we have created well-defined, agreeable tags. Our all tags per image agreement metric shows that the tags perform well in a variety of images, as the average agreement of all the tags is on average over 75\%.

\begin{center} Nose Tags (20 images)

\begin{table}[!htbp]
  \centering
  \begin{minipage}[b]{0.45\linewidth}
    \centering
    \caption{Bridge Projection}
    \begin{tabular}{ccc}
      \hline
      flat & medium & tall \\
      \hline
       0.8409 & 0.7045 & 0.7500  \\
    \end{tabular}
  \end{minipage}
  \begin{minipage}[b]{0.45\linewidth}
    \centering
    \caption{Tip Projection}
    \begin{tabular}{ccc}
      \hline
      flat & medium & tall \\
      \hline
      0.7220   & 0.7272   &0.7500   \\

    \end{tabular}
  \end{minipage}
\end{table}
\begin{table}[!htbp]
  \centering
  \begin{minipage}[b]{0.45\linewidth}
    \centering
    \caption{Nose Width}
    \begin{tabular}{ccc}
      \hline
      narrow & medium & wide \\
      \hline
      0.6250   & 0.7200   & 0.500   \\
    \end{tabular}
  \end{minipage}
  \begin{minipage}[b]{0.45\linewidth}
    \centering
    \caption{Nose Tip Style}
    \begin{tabular}{ccc}
      \hline
      straight & upturned & downturned \\
      \hline
      0.7500 & 0.5500 & 0.7916   \\

    \end{tabular}
  \end{minipage}
\end{table}
\begin{table}[!h]
  \centering
  \begin{minipage}[b]{0.45\linewidth}
    \centering
    \caption{Nose Bridge Style}
    \begin{tabular}{ccc}
      \hline
      upturned & straight & downturned \\
      \hline
      0.8049   & 0.7045   & 0.7500   \\
    \end{tabular}
  \end{minipage}
  \begin{minipage}[b]{0.45\linewidth}
    \centering
    \caption{Nasal Hump}
    \begin{tabular}{cc}
      \hline
       no nasal hump & nasal hump \\
      \hline
      0.9400 & 0.6250 \\

    \end{tabular}
  \end{minipage}
\end{table}
\begin{table}[!ht]
  \centering
  \caption{Nose Image Agreement}
  \begin{tabular}{ccc}
    \hline
    \textbf{highest agreement} & \textbf{lowest agreement} & \textbf{average agreement} \\
    \hline
    1.000 & 0.6667 & 0.800 \\
    \hline 

  \end{tabular}
\end{table}

\vspace{8pt}

\begin{center}Eye Tags (20 Images)
\end{center}

\begin{table}[!ht]
  \centering
  \begin{minipage}[b]{0.45\linewidth}
    \centering
    \caption{Eye Width}
    \begin{tabular}{cccc}
      \hline
      extra small & small & medium & wide\\
      \hline
       0.7500 & 0.6590 & 0.800 & 0.7500  \\
    \end{tabular}
  \end{minipage}
  \begin{minipage}[b]{0.45\linewidth}
    \centering
    \caption{Eye Angle}
    \begin{tabular}{ccc}
      \hline
      upturned & downturned & straight \\
      \hline
      0.6667   & 0.700   & 0.6940   \\

    \end{tabular}
  \end{minipage}
\end{table}
\begin{table}[!ht]
  \centering
  \begin{minipage}[b]{0.45\linewidth}
    \centering
    \caption{Top Crease}
    \begin{tabular}{cc}
      \hline
      top crease & no top crease\\
      \hline
      0.7656 & 0.7140\\
    \end{tabular}
  \end{minipage}
  \begin{minipage}[b]{0.45\linewidth}
    \centering
    \caption{Eye Curvature}
    \begin{tabular}{ccc}
      \hline
      sharp & medium & round \\
      \hline
      0.8210 & 0.7679 & 0.7500   \\

    \end{tabular}
  \end{minipage}
\end{table}

\begin{table}[!ht]
  \centering
  \caption{Eye Image Agreement}

  \begin{tabular}{ccc}
    \hline
    \textbf{highest agreement} & \textbf{lowest agreement} & \textbf{average agreement} \\
    \hline
    1.000 & 0.5625 & 0.7719 \\
    \hline 

  \end{tabular}
\end{table}
\end{center}

\begin{center}Eyebrow Tags (20 Images)

\begin{table}[!ht]

  \centering
  \begin{minipage}[b]{0.45\linewidth}
    \centering
    \caption{Eyebrow Density}
    \begin{tabular}{ccc}
      \hline
      bushy & medium & thin \\
      \hline
       0.6667 & 0.7500 & 0.8210  \\
    \end{tabular}
  \end{minipage}
  \begin{minipage}[b]{0.45\linewidth}
    \centering
    \caption{Eyebrow Thickness}
    \begin{tabular}{ccc}
      \hline
      thin & medium & thick \\
      \hline
      0.7857   & 0.7692   & 0.6667   \\

    \end{tabular}
  \end{minipage}
\end{table}

\begin{table}[!ht]
  \centering
  \begin{minipage}[b]{0.45\linewidth}
    \centering
    \caption{Eyebrow Shape}
    \begin{tabular}{ccc}
      \hline
      s-shaped & round & angled \\
      \hline
      0.7500   & 0.6940   & 0.8330   \\
    \end{tabular}
  \end{minipage}
\end{table}

\begin{table}[!ht]
  \centering
  \caption{Eyebrow Image Agreement}
  \begin{tabular}{ccc}
    \hline
    \textbf{highest agreement} & \textbf{lowest agreement} & \textbf{average agreement} \\
    \hline
    0.9167 & 0.500 & 0.7750 \\
    \hline

  \end{tabular}
\end{table} 

\end{center}

We train two sets of 6 pretrained Resnet152 models on each tag category to predict tags for the nose \cite{pytorchModelsPretrained, He_2016_CVPR}. The first set of models is trained on 568 images from the cleaned fairface sample, annotated manually by researchers. The second set of models is trained on all 2741 images from the clean fairface sample, annotated by Amazon Mechanical Turk workers \cite{mechanicalturk}. The model is evaluated with a 200-image test set hand-annotated by researchers.

\begin{table}[!ht]
  \centering
  \begin{tabular}{lll}
    \hline
    \textbf{tag category} & \textbf{568-image model accuracy} & \textbf{2741-image model accuracy} \\
   \hline
    bridge projection & 0.56 & 0.45 \\

    tip projection & 0.57 & 0.42 \\

    nose width & 0.61 & 0.15 \\

    nose tip style & 0.59 & 0.45 \\

    nose bridge style & 0.55 & 0.56 \\

    nasal hump & 0.95 & 0.66 \\
   \hline
\end{tabular}
\caption{Nose tag model accuracy}
\end{table}

\begin{figure}[!ht]
\begin{minipage}{0.5\textwidth}
    \centering
    \begin{tabular}{lll}
    \hline
    \textbf{tag} & \textbf{precision} & \textbf{recall} \\
    \hline
    width narrow & 0 & 0 \\
    width medium & 0.63 & 1 \\
    width wide & 0 & 0\\
    bridge downturned & 0.15 & 0.13 \\
    bridge upturned & 0.43 & 0.25 \\
    bridge straight & 0.62 & 0.76 \\
    tip downturned & 0.22 & 0.05 \\
    tip upturned & 0.5 & 0.02 \\
    tip straight & 0.63 & 0.95 \\
    no nasal hump & 0.95 & 1 \\
    nasal hump & 0 & 0 \\
    bridge projection flat & 0.58 & 0.61 \\
    bridge projection medium & 0.57 & 0.67 \\
    bridge projection tall & 0 & 0 \\
    tip projection flat & 0.1 & 0.02 \\
    tip projection medium & 0.6 & 1\\
    tip projection tall & 0 & 0 \\
    
    \hline
    \end{tabular}
    \subcaption{568-image model}
\end{minipage}%
\begin{minipage}{0.5\textwidth}
    \centering
    \begin{tabular}{lll}
    \hline
    \textbf{tag} & \textbf{precision} & \textbf{recall} \\
    \hline
        width narrow & 0.12 & 0.19 \\
    width medium & 0.58 & 0.62 \\
    width wide & 0.25 & 0.15\\
    bridge downturned & 0 & 0 \\
    bridge upturned & 0.44 & 0.25 \\
    bridge straight & 0.63 & 0.85 \\
    tip downturned & 0.17 & 0.1 \\
    tip upturned & 0 & 0 \\
    tip straight & 0.61 & 0.85 \\
    no nasal hump & 0.95 & 0.97 \\
    nasal hump & 0 & 0 \\
    bridge projection flat & 0.68 & 0.12 \\
    bridge projection medium & 0.48 & 0.92 \\
    bridge projection tall & 0 & 0 \\
    tip projection flat & 0.15 & 0.08 \\
    tip projection medium & 0.59 & 0.89 \\
    tip projection tall & 0.5 & 0.02 \\
    \hline
    \end{tabular}
    \subcaption{2741-image model}
\end{minipage}
\caption{Precision and recall for all tags, for each model. A score of 0 means there were no predicted instances of the tag.}
\label{fig:pacc}
\end{figure}

The low precision, low recall, and classes not predicted in the test set indicate the model is biased towards the most common classes and as a result is under performing in identifying the other classes (Fig. \ref{fig:pacc}).

\begin{figure}[!h]
    \centering
    \includegraphics[width=\linewidth]{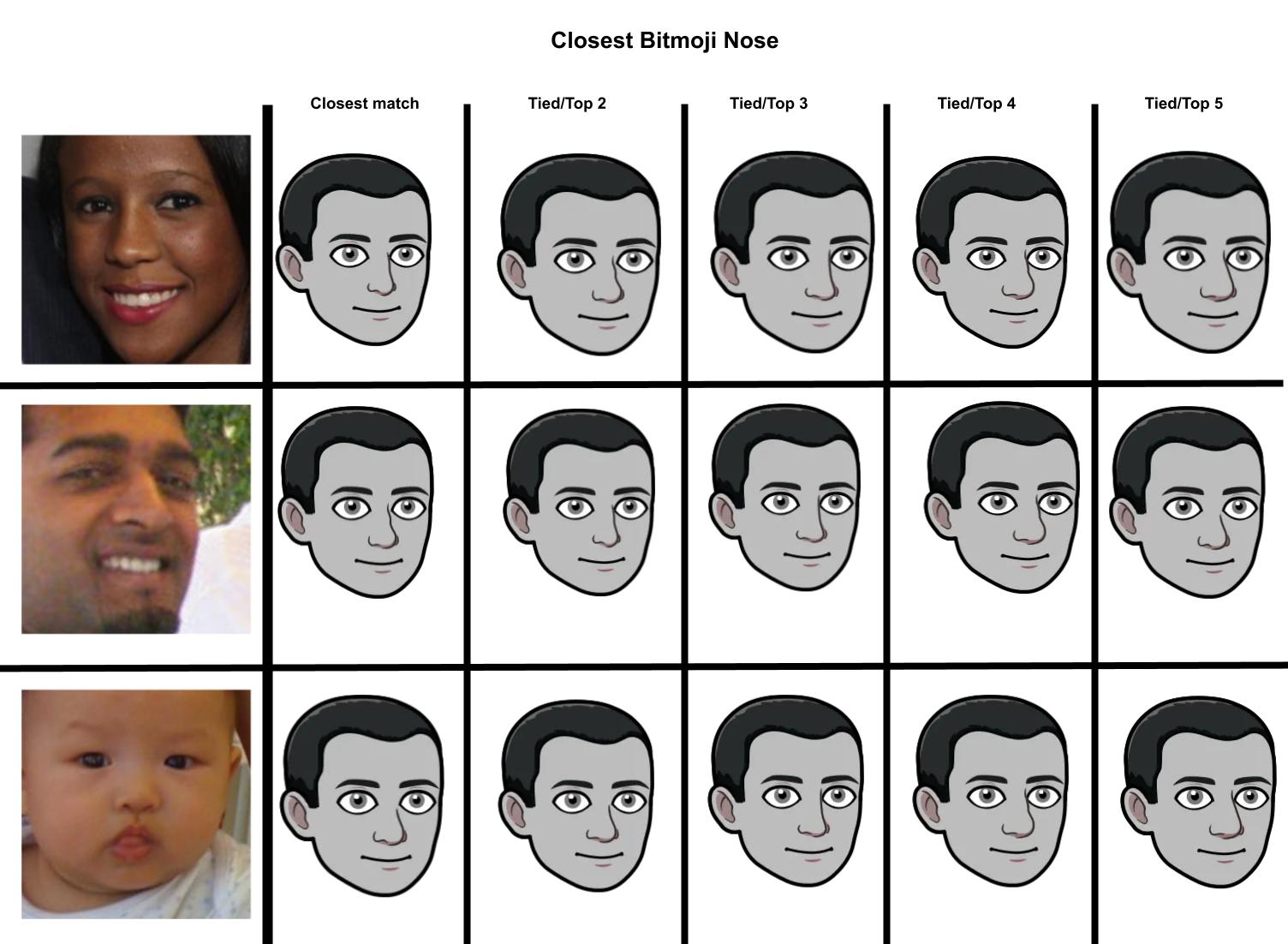}
    \caption{568-image model nose tag prediction converted into a Bitmoji nose using the algorithm described in Sec. 3.7.}
    \label{noses_prediction}
\end{figure}
The translation of tags into Bitmojis exposes further the issue of bias within the model, possibly due to a large class imbalance. The same nose is chosen multiple times, and features seen less in training such as a downturned nose are not detected. 

Some sets of tags point to multiple Bitmoji noses. These are considered ties in our Bitmoji conversion algorithm, yet they can be extremely different from each other. This suggests tag design was not specific enough to isolate nose style (Fig. \ref{tab:distance}).
\section{Conclusion}
We find that although our tags for the nose are agreeable amongst researchers, they do not perform well in a model. We achieve high agreement on most of the nose tags through the use of the reference sheets, a good image sample, and changing tags with annotation metrics in mind. Using this method, we were able to come up with an agreeable set of tags for the nose, eyes, and eyebrows.

\subsection{Limitations}
When exploring points of failure, we have discovered a few limitations involved with tag-based annotation.

\begin{enumerate}
  \item The complexity of the tagging method increases with subtle facial features as the differences become more and more specific to point out.
  \item Some tagging categories can be too heavily influenced by the image taken, which is not consistent.
  \item Class imbalances stemming from tag design limit model performance greatly.
\end{enumerate}

\bibliographystyle{unsrt}  
\bibliography{references}

\end{document}